# A Framework for Exploring Non-Linear Functional Connectivity and Causality in the Human Brain: Mutual Connectivity Analysis (MCA) of Resting-State Functional MRI with Convergent Cross-Mapping and Non-Metric Clustering

Axel Wismüller*[1-4], Xixi Wang[2], Adora M. DSouza[3] and Mahesh B. Nagarajan*[1]

[1]Department of Imaging Sciences, University of Rochester, Rochester, NY, USA

[2]Department of Biomedical Engineering, University of Rochester, Rochester, NY, USA

[3]Department of Electrical Engineering, University of Rochester, Rochester, NY, USA

[4]Faculty of Medicine and Institute of Clinical Radiology, Ludwig Maximilian University, Munich Germany

**Abstract:** We present a computational framework for analysis and visualization of non-linear functional connectivity in the human brain from resting state functional MRI (fMRI) data for purposes of recovering the underlying network community structure and exploring causality between network components. Our proposed methodology of non-linear mutual connectivity analysis (MCA) involves two computational steps. First, the pair-wise cross-prediction performance between resting state fMRI pixel time series within the brain is evaluated. The underlying network structure is subsequently recovered from the affinity matrix constructed through MCA using non-metric network partitioning/clustering with the so-called Louvain method. We demonstrate our methodology in the task of identifying regions of the motor cortex associated with hand movement on resting state fMRI data acquired from eight slice locations in four subjects. For comparison, we also localized regions of the motor cortex through a task-based fMRI sequence involving a finger-tapping stimulus paradigm. Finally, we integrate convergent cross mapping (CCM) into the first step of MCA for investigating causality between regions of the motor cortex. Results regarding causation between regions of the motor cortex revealed a significant directional variability and were not readily interpretable in a consistent manner across all subjects. However, our results on whole-slice fMRI analysis demonstrate that MCA-based model-free recovery of regions associated with the primary motor cortex and supplementary motor area are in close agreement with localization of similar regions achieved with a task-based fMRI acquisition. Thus, we conclude that our computational framework MCA can extract and visualize valuable information concerning the underlying network structure and causation between different regions of the brain in resting state fMRI.

**Key words**: Mutual connectivity analysis, MCA, convergent cross-mapping, CCM, functional MRI, fMRI, functional connectivity, non-metric clustering, resting-state fMRI, Louvain method

*Both authors contributed equally towards the preparation of this manuscript.

## Introduction

We have recently witnessed a significant growth of research endeavors aimed at unraveling characteristic features of structural and functional connectivity in neuro-imaging investigations through functional MRI, e.g. [Marguiles et al. 2010, Li et al. 2009]. We fully expect the insights developed through such scientific efforts to revolutionize our capabilities in further understanding key aspects of brain function. Our specific interest concerns the quantification of cognitive information transfer in regional interactions, which is a challenging methodological problem in the field of computational neuroscience. In this research context, a fundamental problem concerns the accurate analysis of functional/effective connectivity at fine-grained spatial and temporal resolution scales, based on the acquisition capabilities provided by the most advanced contemporary *in vivo* neuro-imaging techniques, such as state-of-the-art functional MRI (fMRI).

Currently, this problem is addressed through a wide range of methods such as seed-based functional connectivity analysis [Biswal et al. 1995, Marguiles et al. 2007, Sun et al. 2005] and its variants, as well as Principal Component Analysis (PCA) [Zhong et al. 2009], Independent Component Analysis (ICA) [Kiviniemi et al. 2003, van de Ven et al. 2004, Beckmann et al. 2005, Damoiseaux et al. 2006], etc. However, many such techniques involve rather inappropriate simplifications that obscure the characteristic properties of the complex system under investigation. For instance, *linear* analytic approaches that are frequently used to characterize brain connectivity can irreversibly discard valuable information; such information could contribute to a more accurate characterization of brain connectivity as a complex *non-linear* system. As another example, Granger causality has been widely used in previous studies involving connectivity and causality analysis in fMRI [Zhou et al. 2009, Deshpande et al. 2012, Stephan et al. 2012, Friston et al. 2013]. However, this approach implicitly introduces the assumption of time-series separability, which is generally unattainable in non-linear dynamic systems observed in nature [Deyle et al. 2011, Takens 1981], and hence can lead to inaccurate or misleading results [Sugihara et al. 2012]. Another drawback with certain approaches, such as region-of-interest seed-based approaches, clustering techniques, PCA, ICA, etc., is that they transform the original high-dimensional imaging data into simpler lower-dimension representations which further limits the interpretation of brain connectivity analysis in physiological and diseased states.

Our primary goal with this contribution is to present a computational framework for exploring functional network connectivity of information transfer in the human brain, while simultaneously circumventing the inevitable information loss induced by the previously mentioned widely used contemporary techniques. The avoidance of initial linearization, dimension reduction or other heuristic simplification steps minimizes the loss of essential information and could allow for accurate recovery of cognitive network structure. To this end, we introduce a mutual connectivity analysis (MCA) approach that can be combined with the application of convergent cross-mapping to non-linear functional connectivity analysis in large time-series ensembles obtained from resting state fMRI neuro-imaging studies. Our approach involves network identification through large scale non-linear mutual time-series prediction followed by functional network identification by partitioning the resulting affinity (or dissimilarity matrix) through non-metric clustering or graph-partitioning methods.

MCA exploits the similarity between time-series as reflected by their mutual non-linear predictability for network identification [Wismüller et al. 2010, Wismüller 2011, Wismüller 2002]. Here, a major difficulty arises from the large number of time-series for analyzing functional connectivity on a state-of-the-art neuro-imaging voxel resolution scale with possible whole-brain coverage, which results in huge dissimilarity matrices $D$. Requiring the appropriate number of nonlinear time-series predictors in order to compute the entire affinity matrix seems to be a tall order, even in this era of super-computing. Here, we present another key contribution by introducing the idea of decoupling the prediction procedure into an initial computationally expensive step of $O(N)$ followed by a computationally inexpensive step of complexity $O(N^2)$, thereby effectively decreasing the empirical complexity of the dissimilarity matrix computation from $O(N^2)$ to $\sim O(N)$. In this work, we investigate the general local model (GLM) as well as a generalized radial basis function (GRBF) neural network for the predictor model. Once the dissimilarity matrix is constructed, we investigate the use of non-metric clustering or graph-partitioning techniques, such as the so-called "Louvain method" [Blondel et al. 2008] for community detection in large networks for identifying specific functional communities in the brain.

We demonstrate our approach in the research context of identifying and visualizing the motor cortex through analysis of time series ensembles in resting-state fMRI data. Previously, Biswal et al. have shown that low-frequency fluctuations (< 0.1 Hz) from regions of the motor cortex associated with hand movement are strongly correlated both within and across hemispheres [Biswal et al. 1995]. Rather than further investigate correlation, our interest lies in exploring non-linear connectivity and causality between time series ensembles from different regions of the motor cortex associated with hand movement, such as the left and right motor cortices (LMC and RMC), and the supplementary motor area (SMA). In the remainder of this paper, we present novel approaches for - (1) recovering the network structure of the motor cortex by computing pair-wise dissimilarity matrices based on cross-prediction of fMRI time-series, (2) exploring and inferring causality between different regions of this network, and (3) concisely visualizing information unveiled by methods (1) and (2) in order to facilitate knowledge discovery in large-scale fMRI connectivity networks.

## Materials and Methods

**Data**: Functional MRI images were acquired from 4 healthy volunteers (1 female and 3 male between 25 to 28 years in age) with a 1.5T GE SIGNA$^{TM}$ whole-body MRI scanner (GE, Milwaukee, WI, USA). Two image sequences were acquired from each subject; the first was under resting state conditions while the second one involved a motor activity task to stimulate the motor cortex. During the resting-state scan, the subject was instructed to stay still and keep eyes closed. For the task sequence, the subject engaged in alternating states of rest and activity for a duration of ~20 seconds each; the activity involved finger-tapping in both hands. The purpose of the motor stimulation experiment in this study was to aid in the localization of motor cortex (left motor cortex - LMC, right motor cortex – RMC, and supplementary motor area - SMA) on the acquired image slices.

Functional MRI (EPI-BOLD) sequences were performed with the following parameters - echo time (TE) - 40 ms, echo-repetition time (TR) - 500 ms, and flip angle (FA) - 90°. The acquisition lasted 4 minutes 16 s during which 512 fMRI scans were acquired from two slice locations that corresponded to the motor cortex. Each image had a slice thickness of 10 mm and an in-plane pixel resolution of 3.75 mm x 3.75 mm (image matrix of 64 x 64). In order to avoid any impact on the data analysis by initial saturation effects [Friston et al. 1994], the first 24 time points (12 seconds) of the acquired fMRI data were discarded.

**Pre-processing**:  Motion artifacts were compensated by automatic image alignment with a widely used registration software package (AIR) [Woods et al. 1992]. To remove effects of signal drifts stemming from the MRI scanner and/or physiological changes in the subjects themselves, linear de-trending was applied; this was determined to be equivalent to high-pass filtering of the time series with a cut-off frequency of 0.0083 Hz. In addition, resting state fMRI time series were subject to low pass filtering with a cut-off frequency of 0.08 Hz for minimizing the influence of respiratory and cardio-vascular oscillations while preserving the frequency spectrum pertaining to functional connectivity [Biswal et al. 1995]. The low-pass filtering was achieved through suppression of appropriate frequency coefficients in the Fourier spectrum, as previously implemented in [Liao et al. 2010]. To aid localization of the motor cortex for purposes of establishing ground truth, spatial smoothening with a Gaussian kernel was applied to the task sequence alone. Finally, the time-courses were further normalized to zero mean and unit standard deviation to focus on signal dynamics rather than amplitude [Wismüller et al. 2002]. After all pre-processing and segmentation of the brain from the skull and background, each slice had ~600-700 pixel time series, and each time series had 488 time points.

**Ground Truth**: The fMRI task sequence was used to establish the ground truth for localization of regions associated with the motor cortex related to hand movement. This was achieved by cross-correlating the task-related stimulus function, i.e., a square waveform corresponding to 20.8 s of rest and 20.8 s of finger-tapping over 6 cycles, with each pixel time series. An empirical threshold of ~0.5-0.6 was used to select pixels whose time series were considered as highly correlated with the stimulus function. A ground truth example in one subject is shown in Figure 1.

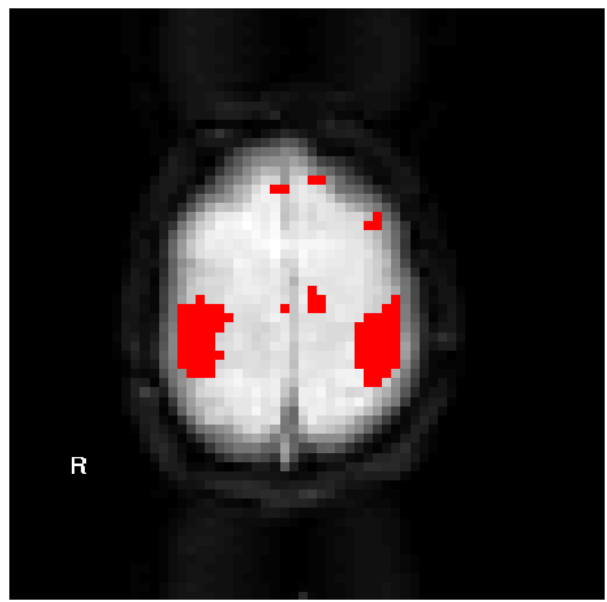

**Figure 1**: The left and right motor cortex (red) associated with finger-tapping, as identified through cross-correlating pixel time series in the fMRI task sequence with the task-related stimulus function. Note that a few pixels corresponding to the supplementary motor area are also captured (mid-line center). A few pixels not associated with the motor cortex are also noted in the anterior right and anterior left frontal cortex, which were identified and excluded for further analysis.

**Mutual Connectivity Analysis (MCA)**: Our first step is to build a pair-wise affinity/similarity matrix $A$ for all time series' under consideration. Depending on the specific focus of the analysis, this may include all time series in a specific region-of-interest, an entire fMRI slice, or even the whole brain. Given $N$ pixels, the temporal dynamics of any two pixels $X$ and $Y$ (where both $X, Y \in \{1, \ldots, N\}$) are represented as a time series of length $L$, i.e., $X(t) = \{X(1), X(2), \dots, X(L)\}$ and $Y(t) = \{Y(1), Y(2), \dots, Y(L)\}$. The goal of mutual connectivity analysis is to establish the degree of dynamic coupling between $X(t)$ and $Y(t)$. This is achieved by quantifying the ability of $X(t)$ to predict $Y(t)$, as discussed below.

We first extract $d$-dimensional vectors that correspond to segments of length $d$ from $X(t)$, i.e.

$$x(t) = \langle X(t), X(t+1), X(t+2), \dots, X\big(t + (d-1)\big) \rangle,$$

for $t = 1$ to $t = L - d$. Every vector from $X(t)$ is then associated with a target vector from $Y(t)$, e.g., a 1-D mapping point on $Y(t)$, i.e., $y(t) = \langle Y(t+d) \rangle$, for $t = 1$ to $t = L - d$. The goal is to then compute an estimate of $Y(t)$, i.e. $\hat{Y}(t)$, from $X(t)$ and evaluate how closely it matches $Y(t)$. We pursue two approaches to accomplish this -

1. General local model (GLM)

For every $x(t)$, we identify $d + 1$ nearest neighbors from $X(t)$. An estimate of the corresponding mapping point on $Y(t)$, i.e. $\hat{y}(t)$, is calculated as the weighted average of the target vectors of these $d + 1$ neighbors. Thus,

$$\hat{y}(t) = \sum_{n=1}^{d+1} w_n y(t_n),$$

where $y(t_n)$ is the target vector of the $n^{\text{th}}$ neighbor on $Y(t)$, and the weight $w_n$ is determined as

$$w_n = \frac{e^{-dist(x(t), x(t_n))/dist(x(t), x(t_1))}}{\sum e^{-dist(x(t), x(t_n))/dist(x(t), x(t_1))}},$$

where $dist()$ represents the Euclidean distance between its arguments. This process is repeated until $\hat{Y}(t)$ is constructed. This local model corresponds to [Sugihara et al. 2012]. Note, however, that a wide range of local linear or average models known from the literature may be used in this step, including those described in [Sugihara et al. 1990, Takens 1981, Farmer 1987].

2. Generalized Radial Basis Function (GRBF) Neural Network

Here, the set of $x(t)$ and their corresponding $y(t)$ are split into a training (Tr) and test (Te) set. The training set is then used to create a non-linear mapping $F$, i.e. $y^{Tr}(t) = F\big(x^{Tr}(t)\big)$. Once defined, this mapping is used to process the set of $x^{Te}(t)$ in the test set, i.e., $\hat{y}^{Te}(t) = F\big(x^{Te}(t)\big)$. These $\hat{y}^{Te}(t)$ are subsequently used to generate $\hat{Y}(t)$.

For defining the approximating function $F$, we use a GRBF neural network with three layers, i.e. the input, hidden and output layers. In the training phase, the activity of neurons in the hidden layer $a_k, k = 1,2, \dots K$, are defined as –

$$a_k(x^{Tr}(t)) = \frac{e^{-\left(x^{Tr}(t) - w_k\right)^2 / 2\rho^2}}{\sum_{j=1}^{K} e^{-\left(x^{Tr}(t) - w_j\right)^2 / 2\rho^2}},$$

for any training set vector $x^{Tr}(t)$ [Moody et al. 1989]. Here, the $\rho$ parameter controls the width of the radial basis function kernel and defines the neighborhood of vectors that contributed to the computation of $F$. The training set $x^{Tr}(t)$ is represented through $K$ prototypical vectors $w_k$, which are computed using an unsupervised clustering approach; fuzzy C-means is used in this study [Bezdek 1981]. In a final step, $\hat{y}^{Tr}(t)$ is computed as a weight sum of the hidden layer activations $a_k$, i.e.,

$$\hat{y}^{Tr}(t) = \sum_{j=1}^{K} a_j(x^{Tr}(t)) \cdot s_j,$$

where $s_j$ are the output weights obtained through minimization of the cost function $E = \|\hat{y}^{Tr}(t) - y^{Tr}(t)\|^2$. After the training phase is completed, $F$ is subsequently used to process the test set. Further details concerning our GRBF approach can be found in [Wismüller 2009].

Regardless of which approach is used, the similarity between $\hat{Y}(t)$ and $Y(t)$ is measured using Pearson's correlation coefficient. Thus, a pair-wise affinity/similarity matrix $A$ for all time series under investigation can be built.

We note that both approaches require processing of $N^2$ pair-wise predictions, which is a computationally expensive task, given that the average number $N$ of pixels per slice to be processed is $N\sim500$-$700$. To address this challenge, we decouple the prediction procedure into a sequence of two steps, namely an initial computationally expensive step of complexity $O(N)$ followed by a subsequent computationally inexpensive step of complexity $O(N^2)$. The computationally expensive steps, i.e. the nearest neighbor search in the GLM approach and vector quantization of the feature vectors in the training set for the GRBF approach, <u>are only performed $N$ times</u>, i.e., only once for each time series. Subsequent steps in both approaches are computationally inexpensive and performed $N^2$ times. Thus, we effectively decreased the empirical complexity of the affinity matrix computation from $O(N^2)$ to $\sim O(N)$.

**Non-Metric Clustering**: From the affinity matrix $A$, we use the Louvain method [Blondel et al. 2008] to recover community structure through non-metric clustering. This approach decomposes a complex network into community clusters with strong intra-community links and weak inter-community links.

Closely related to this algorithm is the concept of modularity, which is a ratio of the density of intra-community node linkage to the density of inter-community node linkage. Modularity $Q$ is defined as

$$Q = \frac{1}{2m} \sum_{i,j} \left[ A_{ij} - \frac{k_i k_j}{2m} \right] \delta(C_i, C_j)$$

where $A_{ij}$ represents the affinity between nodes $i$ and $j$, $k_i = \sum_j A_{ij}$ is the sum of affinities of nodes attached to $i$, $C_i$ is the community to which node $i$ is assigned, $\delta(u, v) = 1$ when $u = v$, and 0 otherwise, and $m = \frac{1}{2} \sum_{ij} A_{ij}$ [Newman 2004]. The Louvain method aims to find high modularity clusters in networks and also establishes a hierarchical community structure for the network to enable different resolutions of community detection [Blondel et al. 2008]. This approach is particularly advantageous for our study, as it does not require a symmetrized affinity matrix, i.e. node $i$ may exhibit a higher affinity toward node $j$ than vice versa.

The algorithm involves an iterative process during which different nodes of the network are merged into larger communities if the modularity is improved as a consequence. The change in modularity $\Delta Q$ achieved when an isolated node $i$ is merged with a community $C$ is calculated as -

$$\Delta Q = \left[\frac{\Sigma_{in}+k_{i,in}}{2m} - \left(\frac{\Sigma_{tot}+k_i}{2m}\right)^2\right] - \left[\frac{\Sigma_{in}}{2m} - \left(\frac{\Sigma_{tot}}{2m}\right)^2 - \left(\frac{k_i}{2m}\right)^2\right],$$

where $\Sigma_{in}$ is the sum of affinities of the links inside $C$, $\Sigma_{tot}$ is the sum of affinities of links incident to nodes in $C$, $k_i$ is the sum of affinities of links incident to node $i$, $k_{i,in}$ is the sum of affinities of the links from node $i$ to $C$, and $m$ is the sum of affinities of all links in the network [Blondel et al. 2008]. The iterative process is continued until no further improvement in modularity can be achieved. Further details pertaining to this clustering approach can be found in [Blondel et al. 2008].

In order to avoid the creation of large super-communities that encompassed smaller and more interesting clusters, we also pursue an approach frequently applied in spectral clustering to make the affinity matrix sparser [Luxburg 2006]. Specifically, we only consider the $k$ most similar nodes for any given node $i$. Additionally, only mutual $k$ most-similar nodes are considered, i.e., bi-directional links in the $k$ most-similar nodes. In this study, $k = 100$ is chosen empirically from preliminary analysis; this corresponds to approximately 20% of the nodes in the network in most cases. Similarity between clustering results and the ground truth was evaluated using the Dice coefficient [Dice 1945], as described in [Wismüller et al. 2010]. When required, different clusters are merged in an iterative process to maximize the dice coefficient achieved with the ground truth.

**Causality Analysis**: We further extend MCA to investigate causality between different regions in the brain. Previously presented in the literature as convergent cross-mapping (CCM) [Sugihara et al. 2012], this process evaluates causation in dynamic systems through an examination of cross-prediction between two series (say $X(t)$ and $Y(t)$). A key feature of CCM is its exploration of convergence, a phenomenon where the ability of time series $X(t)$ to predict (or "cross-map") $Y(t)$ improves as the time series length $L$ increases. The length $L$ is modified in this study by taking varying percentages of the time series for MCA with GLM or by using varying percentages of the time series as the training set for MCA with a GRBF neural network (10% - 80% in both cases). Thus, by observing the degree to time series $X(t)$ and $Y(t)$ are cross-mapped over increasing $L$, one can establish grounds for causation between both time series (if any causation exists in the first place).

In this study, we restrict the examination of causation to specific regions of the motor cortex, as identified using the ground truth. Thus, MCA with both GLM and GRBF is used to build a smaller affinity matrix involving pixels from the LMC, RMC and SMA alone. From the entire collection of vectors $x(t)$ from time series $X(t)$, a randomly chosen subset (of 10-80%) can have different variations. So we compute an affinity matrix for 20 different variations of training subsets of $x(t)$, and use their average for CCM analysis, as seen in Figure 2 (top row).

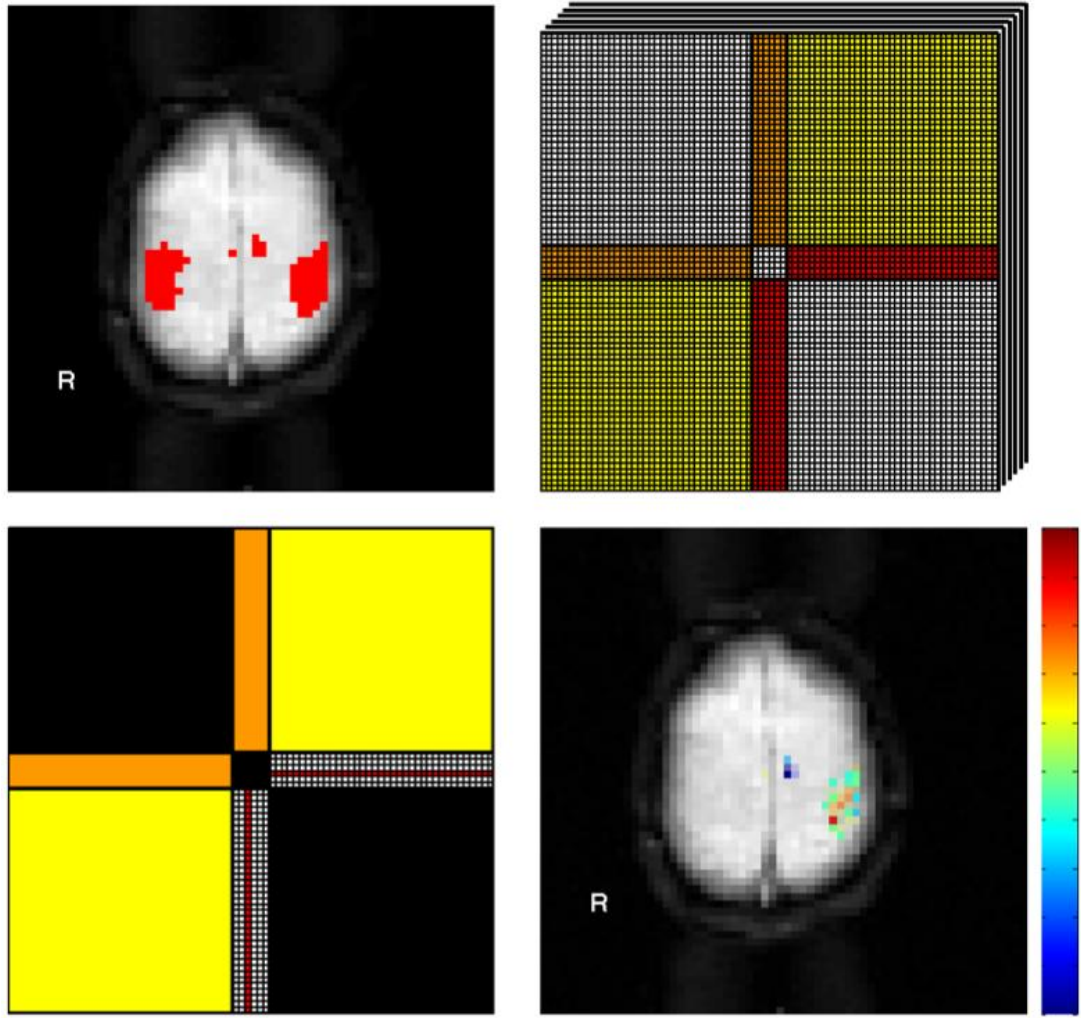

**Figure 2**: Overview of causality analysis and visualization. (TOP LEFT) Ground truth; pixels colored RED are indicative of regions of the primary sensory motor cortex associated with hand movement. (TOP RIGHT) Affinity matrix, as computed with MCA. Note that the areas colored in red, orange and yellow correspond to cross-prediction between different regions, which is our focus in this study. Note that the stacked appearance of this matrix indicates that multiple such matrices are computed. (BOTTOM LEFT) The averaged affinity matrix. A specific pixel from the SMA and its cross-mapping with the LMC is shown; its corresponding matrix entries are marked in red. (BOTTOM RIGHT) Influence scores for all pixels in the supplementary motor area and the left motor cortex based on cross-mapping between both regions. Blue and red coloring indicate unidirectional influence from and to pixels in the other region respectively.

Rather than restrict ourselves to a purely pixel-wise analysis of causality, we also perform a global causality analysis by treating each of the three regions of the motor cortex as individual nodes. Computationally, this is achieved by averaging within the six colored regions of the affinity matrix shown in Figure 2 (top right).

**Visualization**: To better understand the results achieved with CCM analysis, we utilize a specific visualization procedure. When comparing any two regions of the motor cortex, each pixel is assigned an influence score which takes into account - (1) how well the time series of pixels in region #2 are predicted by the pixel time series in region #1 and (2) how well pixel time series in region #2 are able to predict the time series of the pixel in region #1. Thus, for a pixel $i$ in region #1, the influence score $I_i$ is computed as -

$$I_i = \sum_j A_{ij} - \sum_j A_{ji},$$

for $j$ pixels in region #2. These influence scores are color coded and overlaid on the image, as illustrated in Figure 2 (bottom right).

All procedures were implemented using MATLAB 8.1 (MathWorks Inc., Natick, MA, 2013). The Louvain method implementation was taken from [Scherrer 2008].

## Results

### 1. Recovery of motor cortex community structure from MCA

Figure 3 shows the results of recovering communities associated with the motor cortex from a single resting state fMRI slice through non-metric clustering with the Louvain method. As seen here, non-metric clustering with the Louvain method is able to recover the community structure of the motor cortex from resting state fMRI time series ensembles.

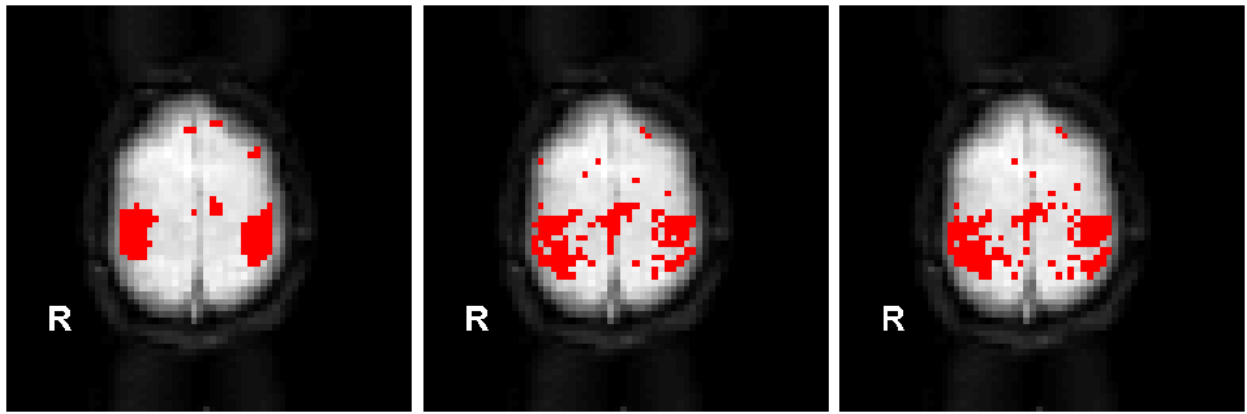

**Figure 3**: (LEFT) Ground truth for subject #1; red pixels show activations in regions of the primary sensory motor cortex associated with finger tapping. (MIDDLE) Clustering results from MCA with GLM; Dice coefficient with ground truth is 0.51. (RIGHT) Clustering results from MCA with GRBF neural network, Dice coefficient with ground truth is 0.50. As seen here, regardless of the MCA approach used to build the affinity matrix, one can recover the community structure of the motor cortex through non-metric clustering with the Louvain method.

Table 1 shows the Dice coefficients obtained from the clustering results of the motor cortex with the ground truth on different slices acquired from different subjects. As seen here, no distinct differences were noted between the clustering results achieved with MCA between the GLM and GRBF network approaches. However, we do note that the regions of the motor cortex, as captured through clustering, appear to be larger in size when compared to the ground truth, hence explaining the poorer magnitudes of the Dice coefficient.

| Subject # | Slice # | Dice Coefficient | |
|:---:|:---:|:---:|:---:|
| | | MCA with GLM | MCA with GRBF |
| 1 | 1 | 0.41 | 0.45 |
| | 2 | 0.51 | 0.50 |
| 2 | 1 | 0.37 | 0.34 |
| | 2 | 0.31 | 0.32 |
| 3 | 1 | 0.54 | 0.50 |
| | 2 | 0.45 | 0.49 |
| 4 | 1 | 0.33 | 0.37 |
| | 2 | 0.37 | 0.34 |

**Table 1**: Dice coefficients between the motor cortex regions identified through clustering and the ground truth obtained from the fMRI task sequence. Results are reported for two slices on all four subjects who participated in this study. As seen here, no distinct differences were noted between clustering results achieved by MCA with GLM and GRBF approaches.

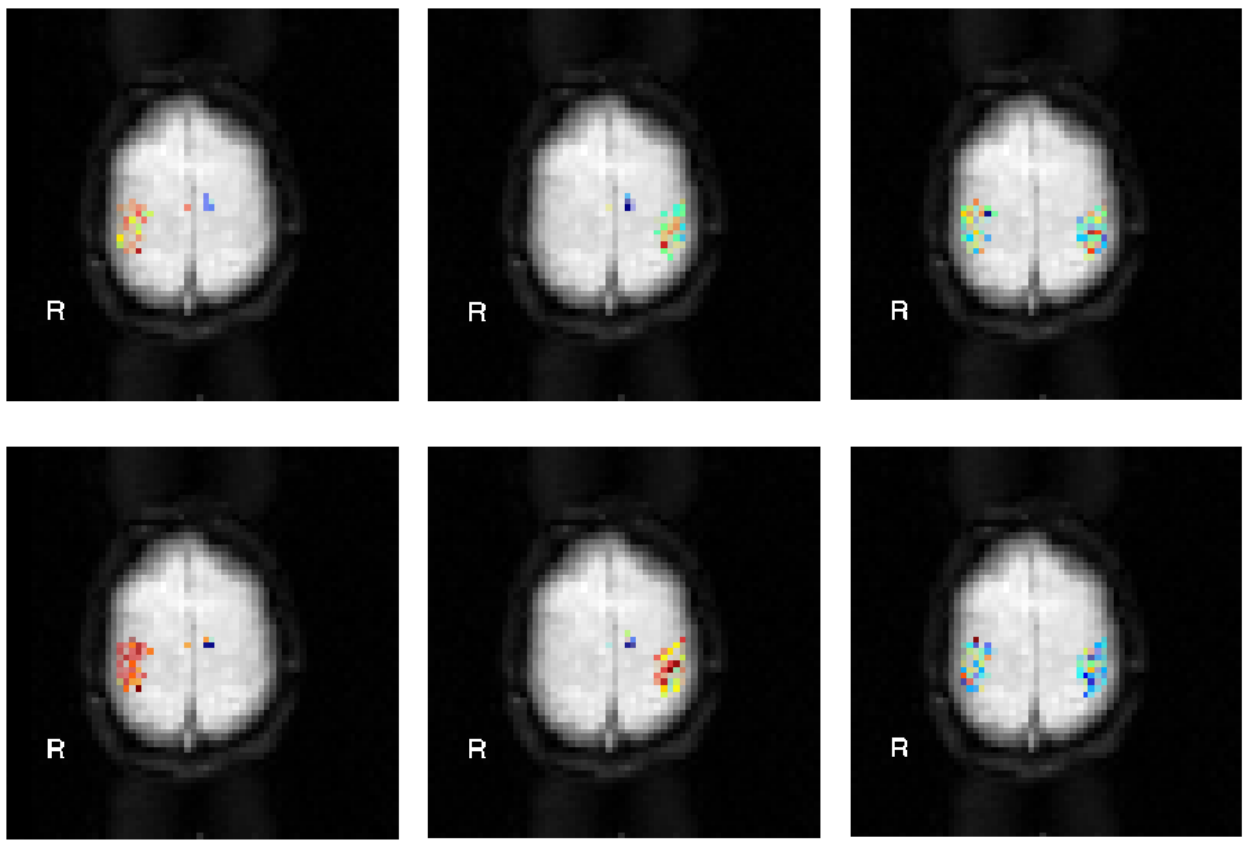

**Figure 4**: (TOP ROW) Pair-wise regional causality analysis with MCA+GRBF performed on a pixel-wise basis as outlined in Figure 2. (BOTTOM ROW) Pair-wise regional causality analysis with MCA+GLM performed on a pixel-wise basis as outlined in Figure 2. From left to right, the pair-wise comparisons involve the RMC and SMA, LMC and SMA, and the LMC and RMC. Note that a 60% subset of the original feature vectors was used for generating these results.

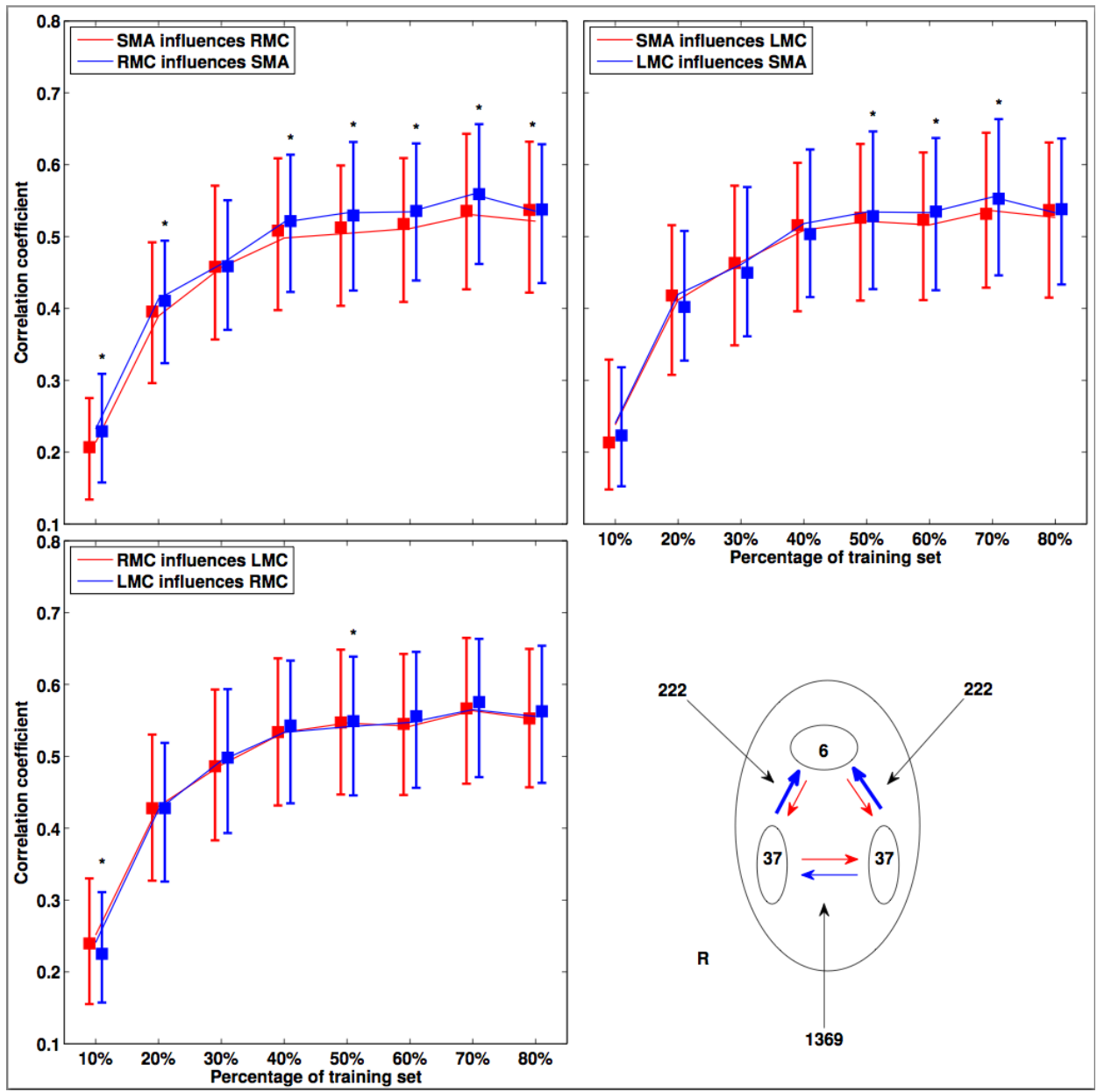

**Figure 5**: Overview of global causality analysis with CCM+GRBF. (TOP LEFT) A comparison of cross-prediction performance (or quality of cross-mapping estimates) between the supplementary motor area and the left motor cortex as a function of the training subset percentage used in CCM analysis. The slightly higher correlation values indicated by the blue curve suggest asymmetric influences between these two regions; specifically, the left motor cortex has a stronger influence on the supplementary motor area than vice versa. Note that every point on these curves is represented as a distribution where the central marker is the median and the edges are the 25th ad 75th percentile respectively, since the cross-prediction performance is repeated 20 times. Significant differences ($p <$ 0.05) between distributions from different curves for the same percentage of the training set used are marked with an asterisk. (TOP RIGHT) A comparison of cross-prediction performance between the supplementary motor area and the right motor cortex as a function of the training subset percentage used in CCM analysis. Again, the right motor cortex appears to have a stronger influence on the supplementary motor area than vice versa. (BOTTOM LEFT) A comparison of cross-prediction performance between the left and right motor cortices as a function of the training subset percentage used in CCM analysis. Here, the high Pearson correlation coefficient values suggest a significant bi-directional coupling between the two regions; however, this coupling does not exhibit signs of asymmetric influence in a particular direction. (BOTTOM RIGHT) Summary of causality between the three regions. The thicker arrows between regions indicate asymmetric bi-directional coupling between two regions, i.e. the influence of one region on the other is stronger than vice versa. The number of pixel-wise comparisons for establishing causality between a pair of regions is also specified.

## 2. Pixel-wise causality analysis

Figure 4 shows a visualization of the results of pair-wise causality analysis between the LMC, RMC and SMA for the same image slice in Figure 3 using our visualization approach (illustrated in Figure 2). As seen here, similar results are observed for both MCA models, i.e. GLM and GRBF network. A specific direction of causation is noted between the different regions in Figure 3. Specifically, both LMC and RMC appear to influence the SMA. However, we also note that such findings are not consistent on images from other subjects.

## 3. Global causality analysis

By treating each region as an individual node, the causality analysis is simplified and allows for a simpler illustration of cross-prediction performance between different regions as a function of the time series length $L$, or in our study, the training sub-set percentage (10% - 80%). Differences in these cross-prediction curves for two regions indicate an asymmetric directional influence in their bi-directional coupling. These results are shown in Figure 5 for MCA with GRBF, as performed on the same image slice shown in Figure 3. As seen here, the results are in close agreement with the pixel-wise causality analysis results shown in Figure 4. However, the directions noted here are not consistent across different subjects.

# Discussion

We present a framework for analysis of functional connectivity in the brain from resting state fMRI data for purposes of recovering the underlying network structure and establishing causality. While other methodologies for assessing functional connectivity through fMRI exist, such as seed-based approaches [Biswal et al. 1995, Marguiles et al. 2007, Sun et al. 2005, Zhong et al. 2009], ICA [Kiviniemi et al. 2003, van de Ven et al. 2004, Beckmann et al. 2005, Damoiseaux et al. 2006] etc, our framework avoids certain shortcomings such as assumptions of linearity, time-series separability, etc which may result in discarding valuable information and may even lead to misleading results or inaccurate conclusions. We instead propose to use non-linear mutual connectivity analysis (MCA) to evaluate the pair-wise cross-prediction quality between resting state fMRI time series acquired from the brain. Our proposed methodology also takes into account the rather large number of time series to be analyzed, given the resolution capabilities of state-of-the-art fMRI and presents innovative strategies to make the computational load more manageable. Our results suggest that such pair-wise affinity matrices can not only be computed with manageable computational effort, but can also reveal valuable information concerning the underlying network structure and causation between such regions.

We demonstrate the use of MCA with two approaches, i.e. GLM and GRBF networks, for building an affinity matrix that stores the cross-prediction quality between pairs of time series. The affinity matrix is subsequently subject to non-metric clustering with the Louvain method for purposes of identifying regions associated with the primary sensory motor cortex. Our results, as seen in Figure 3, show that the motor cortex network is recovered regardless of which MCA approach is used. We do note that the LMC, RMC and SMA regions detected as part of the motor cortex on resting-state fMRI using our model-free approach appear larger than on the task sequence. This is likely due to the fact that only those portions of the motor cortex associated with the finger-tapping task performed by the subjects are detected on the task fMRI sequence. Thus, it is possible that our model-free approach with non-metric clustering on resting state fMRI provides a more complete identification of the motor cortex itself. However, any further localization of regions within the motor cortex with respect to specific organs may not be possible without a stimulus task; a hypothesis that will be investigated further in future studies.

We also note that other methods of non-metric clustering such as agglomerative clustering [Duda et al. 2001], pair-wise clustering through deterministic annealing [Hofmann et al. 1997], topographic mapping of proximity data (TMP) [Graepel et al. 1998], spectral clustering methods based on eigenvalue decompositions of graph Laplacians [Shi et al. 2000, Ng et al. 2001] etc., can be used in place of the Louvain method, as seen in Figure 6. However, some of these methods require a symmetrized affinity matrix, which could constitute an information loss that adversely affects network structure recovery. One exception is spectral clustering, which has been shown to extend to directed graphs [Meila et al. 2007].

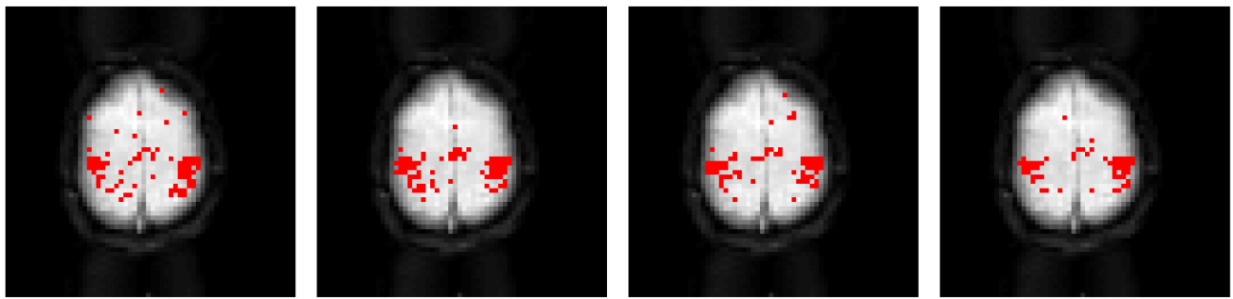

**Figure 6**: Motor cortex network recovery with different non-metric clustering methods. From left to right - agglomerative clustering, temporal mapping of proximity data, Louvain method and spectral clustering. As seen here, all clustering methods are able to recover the LMC, RMC and SMA regions of the motor cortex network. Note that a comprehensive analysis of different non-metric clustering methods is considered out of scope of this contribution, and will be explored in a different study.

While it is clear that the LMC, RMC and SMA regions of the motor cortex can be recovered using MCA, the results of causality analysis between these regions is not quite so straightforward to interpret. Our results do not reveal a consistent direction of causation between these regions across different subjects. However, as noted in Figure 5, an exploration of cross-prediction performance as a function of time series length suggests some form of bi-direction coupling between these regions. Furthermore, an evaluation of pixel-wise causality analysis results across different subjects suggests that all pixels within the same region may not behave homogeneously as a single unit. Thus, our global causality analysis approach may, in some cases, oversimplify the true picture of causation within nodes of the motor cortex network. However, we feel that our visualization procedures for both pixel-wise and global causality are particularly useful tools that complement our MCA approach and enable straightforward analysis of the results achieved, and can be easily extended for analysis of other networks such as the default mode network, executive control network, auditory network etc.

We acknowledge the use of low pass filtering at a cut-off frequency of 0.08 Hz, which is now considered to be a rather controversial step in fMRI data post-processing. While several studies in the literature have used this step to eliminate the influence of respiratory and cardio-vascular oscillations while preserving the frequency spectrum pertaining to functional connectivity [Biswal et al. 1995, Liao et al. 2010], arguments have been made that such low-pass filtering can artificially induce correlation in fMRI resting state time series data [Davey et al. 2013] or erroneously discard high frequency components. While we have nothing new to add to this controversy at this point, we would like to investigate the impact of low pass filtering on the results of connectivity and causality analysis achieved with our methodology in future studies. In the absence of low pass filtering, we expect that MCA with the GRBF model to be more robust to the higher frequency components of resting state fMRI time series data than the GLM model. We also anticipate that our strategy to make the affinity matrix sparser may require some more fine-tuning to work with unfiltered resting state fMRI time series data.

To the best of our knowledge, this is the first publication in the literature, which introduces convergent cross-mapping (CCM) into the domain of fMRI analysis. Here, it should be noted that mutual connectivity analysis (MCA) [Wismüller 2011, Wismüller et al. 2010, Wismüller 2002] and CCM [Sugihara et al. 2012] complement each other in the sense that both techniques can be interpreted as extensions of the other. If the mutual non-linear predictability is evaluated for a given time-series training set, CCM is computationally equivalent to the first step of MCA. By simply repeating this predictability evaluation for different amounts of time-series training data, the procedure can be re-interpreted as CCM, which according to [Sugihara et al. 2012] also allows the user to infer a directed causal influence between the time-series under investigation. In this sense, CCM is an extension of the first step of MCA. However, CCM does not include the second step of MCA, namely the idea of community detection with non-metric clustering or graph-partitioning methods, based on the connectivity structure revealed in the preceding mutual predictability evaluation step. In addition, CCM can also be seen as a special case of MCA in the sense that MCA eliminates the restriction to local simplex models with their inherent leave-one-out cross-validation approach used in [Sugihara et al. 2012] to a much broader scope of both non-linear time-series prediction models and cross-validation schemes, which makes it applicable in a generic machine learning context. This is demonstrated by using a neural network based non-linear time-series predictor, namely the GRBF neural network approach used both here and in our previous publications [Wismüller 2011, Wismüller et al. 2010, Wismüller 2002]. Further generalization to other non-linear function approximators, such as support vector techniques, random forests, etc. appears as straightforward extension of our work.

Although MCA and CCM have been developed and published independently at different times and in different application contexts, both methods complement each other in the sense that their combination constitutes a powerful novel approach to analyzing functional connectivity and causation in complex systems. Beyond its immediate neuro-imaging applications in computational neuroscience, neurophysiology, and clinical neurology, we conjecture that our approach will be useful in many other research domains throughout science and engineering, ranging from information retrieval to systems biology.

## Conclusion

We present a framework for analysis of functional connectivity and causality in the brain from resting state fMRI data, using MCA with GLM and GRBF approaches to quantify pair-wise affinity between different pixel time series and subsequent non-metric clustering with the Louvain method, for recovering any underlying network structure. The results observed in our study suggest that our methodology constitutes a model-free approach to recovering the network structure of the motor cortex. However, we do not note any consistent direction of causation between the individual regions of the motor cortex, i.e. the left and right primary motor cortices, and the supplementary motor areas, across subjects. The methodology for analysis and visualization of functional connectivity and causality presented in this contribution could easily be extended to the whole brain for analysis of different networks. Future applications of our method could include quantification of functional connectivity under pathological conditions for purposes of establishing imaging biomarkers of diseased states and therapeutic effectiveness. However, controlled trials with larger human subject cohorts need to be conducted in order to further validate the applicability of our approach in a clinical setting.

## Acknowledgements

This research was funded by the National Institutes of Health (NIH) Award R01-DA-034977. The content is solely the responsibility of the authors and does not necessarily represent the official views of the National Institute of Health. The authors would like to thank Dr. Lutz Leistritz and Prof. Dr. Herbert Witte of Bernstein Group for Computational Neuroscience Jena, Institute of Medical Statistics, Computer Sciences, and Documentation, Jena University Hospital, Friedrich Schiller University Jena, Germany, Dr. Oliver Lange and Prof. Dr. Maximilian F. Reiser of the Institute of Clinical Radiology, Ludwig Maximilian University, Munich, Germany and Prof. Dr. Dorothee Auer of Institute of Neuroscience, University of Nottingham, UK, for their support.